\documentclass{Interspeech}

\interspeechcameraready

\title{Assessment of L2 Oral Proficiency using Speech Large Language Models}

\author[affiliation={}]{Rao}{Ma}
\author[affiliation={}]{Mengjie}{Qian}
\author[affiliation={}]{Siyuan}{Tang}
\author[affiliation={}]{Stefano}{Bannò}
\author[affiliation={}]{Kate M.}{Knill}
\author[affiliation={}]{Mark J.F.}{Gales}

\affiliation{ALTA Institute, Department of Engineering}{University of Cambridge}{UK}
\email{\{rm2114,mq227,st941,sb2549,kmk1001,mjfg100\}@cam.ac.uk}

\usepackage{comment}

\usepackage[table]{xcolor}
\usepackage{amsmath}
\usepackage{graphicx}
\usepackage[colorinlistoftodos]{todonotes}
\usepackage{float}
\usepackage{subcaption}
\usepackage{amsfonts}
\usepackage{multicol,multirow}
\usepackage[bottom]{footmisc}
\usepackage{booktabs, pifont}
\usepackage{enumitem}
\usepackage{soul}
\usepackage{tabularx, url}
\definecolor{gray}{rgb}{0.5,0.5,0.5}
\definecolor{Gray}{gray}{0.9}

\usepackage{hyperref}

\begin{document}

\maketitle

\footnotetext[1]{This paper reports on research supported by Cambridge University Press \& Assessment, a department of The Chancellor, Masters, and Scholars of the University of Cambridge.}

\begin{abstract}
The growing population of L2 English speakers has increased the demand for developing automatic graders for spoken language assessment (SLA). Historically, statistical models, text encoders, and self-supervised speech models have been utilised for this task. However, cascaded systems suffer from the loss of information, while E2E graders also have limitations. With the recent advancements of multi-modal large language models (LLMs), we aim to explore their potential as L2 oral proficiency graders and overcome these issues. In this work, we compare various training strategies using regression and classification targets. Our results show that speech LLMs outperform all previous competitive baselines, achieving superior performance on two datasets. Furthermore, the trained grader demonstrates strong generalisation capabilities in the cross-part or cross-task evaluation, facilitated by the audio understanding knowledge acquired during LLM pre-training.
\end{abstract}

\section{Introduction}

With the growing number of second language (L2) English learners worldwide, the demand for building automated systems capable of assessing their spoken language proficiency has been steadily increasing. These systems play a crucial role in scoring the candidates' responses and providing feedback in language learning. Compared to human annotators, building automated graders for spoken language assessment (SLA) has several advantages. Firstly, they are more cost-effective and can be deployed in a large scale once trained, alleviating the expenses associated with hiring and training new human graders. Additionally, employing a system can enhance scoring consistency by reducing the variations and biases in human judgments. Furthermore, the automated system can operate in a timely manner, making it possible to provide real-time feedback to L2 learners to improve their spoken skills.

Throughout the history of developing graders for speaking assessment, various approaches have been explored, with the first systems investigated in the 1990s using statistical models mainly targeting pronunciation of read speech~\cite{bernstein1990automatic, cucchiarini1997automatic, franco2000sri} and only later spontaneous speech~\cite{zechner2007speechrater}. In recent years, deep neural network approaches have brought significant improvements~\cite{qian2012use, evanini2018improvements,qian2019slate}. Another crucial advancement was brought by the application of word embeddings using foundation models, such as BERT~\cite{devlin2019}, on automatic assessment tasks~\cite{raina2020universal, wang2021}. More recently, the use of embeddings from speech foundation models, such as wav2vec 2.0~\cite{baevski} and HuBERT~\cite{Hsu2021HuBERTSS}, has been investigated for mispronunciation detection and diagnosis~\cite{peng21, wu21, xu21} and automatic pronunciation assessment~\cite{eesung}. However, the use of text foundation models such as BERT comes with limitations. First, these models rely on automatic transcription, which introduces the risk of propagating transcription errors into the assessment pipeline, hence adding noise to the evaluation process. Secondly, they cannot explicitly capture acoustic-related information, limiting their ability to assess pronunciation and fluency aspects directly
To address these limitations, \cite{banno2023proficiency, bannoassessment} proposed building an E2E grader from the wav2vec 2.0 model. 
The speech foundations models are capable of leveraging rich acoustic features, however, they do not explicitly account for aspects related to content, vocabulary, grammar, and discourse coherence, which are crucial for a comprehensive speaking assessment. 
The model shows limited performance on some test parts while improved performance can be achieved when further combined with a BERT-based model and a feature-based grader through linear regression. 
\cite{mcknight23_slate} further enhanced the wav2vec 2.0 grader by replacing mean pooling with an attention mechanism, which more effectively aggregates information from the input audio segments, leading to improved performance.


In recent years, there have been significant breakthroughs in developing large language models (LLMs), which have greatly advanced the field of natural language processing. Given their remarkable performance in text processing, researchers have been focusing on extending LLMs' capabilities to handle other modalities, such as the speech input. 
These models are in general built from pre-trained speech encoders and LLMs, which are combined with projection layers for dimension and modality match. After learning from a vast amount of speech data and refining through instruction tuning, models like SALMONN \cite{tangsalmonn} and Qwen-Audio \cite{chu2023qwen, chu2024qwen2} demonstrate strong performance across various tasks such as speech recognition and translation, audio captioning, and spoken question answering. While LLMs have been used extensively for L2 writing assessment~\cite{mizumoto2023gpt, yancey-etal-2023-rating, banno-etal-2024-gpt}, their application to speaking assessment remains largely unexplored. Recently, \cite{fu2024pronunciation} developed a speech LLM for L2 assessment and achieved competitive performance, but only in the limited scope of pronunciation scoring, while the use of speech LLMs for holistic assessment (which therefore includes aspects such as grammar, vocabulary, coherence and cohesion, fluency, etc.) remains unexplored.

Although speech LLMs have achieved state-of-the-art performance on a range of audio understanding tasks, can they be used as L2 proficiency graders and how good are they compared to previous approaches? The SLA task presents a challenge for the model, as not only does it require the comprehension of the spoken content but also the ability to capture rich information from the speech such as pronunciation, speech rate, and prosody. Additionally, the fact that the test takers are L2 speakers add to the complexity of audio understanding, as the model needs to handle diverse accents, speech disfluencies, and potential grammatical errors. Lastly, the limited availability of data due to copyright restrictions, along with the inconsistency in human labelling makes the model development more challenging.
In this paper, we investigate the potential of deploying speech LLMs for accurately assessing L2 oral holistic proficiency. Our study explores several training strategies and decoding approaches for this task on two standard datasets. 


\section{Methodology}


For the spoken language assessment datasets studied in this paper,
reference scores are annotated by human graders on a scale from 1 to 6 for each audio response, with specific instructions provided for each class level. Half-point scores (e.g., 3.5) are allowed for intermediate ratings. Our goal is to develop auto graders that produce scores closely aligned with the references.
As LLMs are trained with the next-word prediction target, they are suitable for tasks with text-based outputs. Thus, applying LLMs to holistic scoring, which requires numerical predictions, presents a unique challenge. 
In the following, we explore several training schemes to address this issue.

\subsection{Training}

\subsubsection{Classification with Cross Entropy Loss (CE)}
Pre-trained on vast amounts of text corpora through next-word predictions, LLMs have shown superior performance as classifiers in tasks such as question answering and sentiment classification \cite{wei2022emergent}. 
Following these practices, we frame L2 grading as an audio classification task, where the model predicts a proficiency level $c$ from a set of predefined classes $C$ (\textit{``A''} to \textit{``F''}) based on the candidate's speech responses. Here, a function \texttt{score} is used to map each text-based grade into its numeric class (e.g. label \textit{``A''} to 6 and \textit{``F''} to 1, etc). Suppose the reference score is $\hat{y}_i$, then $\hat{y}_{i,c}=1$ if $\text{score}(c)=\hat{y}_i$, otherwise it is set to 0. Denote the predicted probability from the model for class $c$ as $y_{i,c}$, we aim to minimise the cross-entropy loss,
\begin{equation}
    \sum_{c=1}^{C} y_{i,c} \log(\hat{y}_{i,c})
\end{equation}
By framing holistic scoring as a text-based classification task, we are able to evaluate the zero-shot performance of the speech LLM for this task, referred to as $\tt zs$ in our experiments.

\subsubsection{Classification with Fair Average Loss (FA)}
For holistic scoring, human annotators are allowed to assign mid-class scores that fall between the descriptors of consecutive levels (e.g. score 5.5 falls between level \textit{``A''} and \textit{``B''}), making them difficult to map and represent with a discrete class label. In the \textit{CE} approach, scores are rounded to the nearest integer, resulting in some loss of information during training and evaluation.
To address this, we propose to represent the model outputs as the fair average of predictions and train the system to minimize the mean squared error (MSE) with the reference score,
\begin{equation}
    \left( \sum_{c=1}^{C} s_{i,c} \cdot \text{score}(c) - \hat{y}_i \right)^2
\end{equation}
Here, \( s_{i,c} \) represents the values derived by applying the softmax function to the model's predictions across all predefined class labels, i.e.,
\begin{equation}
    s_{i,c} = \frac{\exp(y_{i,c})}{\sum_{c=1}^C \exp(y_{i,c})}
    \label{softmax}
\end{equation}

\subsubsection{Regression-based Approach (Reg)}
The holistic scoring task is traditionally viewed as a regression problem, where the model learns to predict a numeric value that minimizes the MSE loss with the reference score $\hat{y}_i \in [1, 6]$. In the \textit{Reg} approach, we add a linear layer on top of the last LLM hidden layer to map the model representation to a scalar value $y_{i}$, with the loss computed as $\left(y_{i} - \hat{y}_i \right)^2$. This approach implicitly encodes the ranking of different scores. Nevertheless, it adds new parameters to the LLM, disabling the zero-shot evaluation. Moreover, although the regression approach is straightforward, it is not fully aligned with the pre-training objectives of LLMs, which are designed to generate text tokens. 

\subsection{Evaluation}

For approaches that view holistic scoring as a classification task, multiple decoding schemes can be utilised in decoding. With greedy decoding, we select the top predicted class and retrieve its associated score as the final prediction, i.e. 
\begin{equation}
    \text{score}(\arg\max_c y_{i,c})
\end{equation}
This approach is referred to as the \textit{``hard''} decoding strategy. Nevertheless, it only takes into account the highest prediction and ignores the model predictions over other class labels. Alternatively, we propose a \textit{``soft''} decoding strategy where we aggregate the model outputs using the fair average calculation. 
\begin{equation}
    \sum_{c=1}^{C}  \left[s_{i,c} \cdot \text{score}(c) \right]
\end{equation}

\subsection{Prompt for Speech LLM}
The following prompt is given to the speech LLMs for training and evaluation, which is based on the CEFR descriptors \cite{council2001common} of different language proficiency levels: \textit{``Predict the overall score of the speech using the options provided below.\textbackslash n\textbackslash nOption A: Can produce clear, smoothly flowing, well-structured discourse with an effective logical structure which helps the recipient to notice and remember significant points.\textbackslash nOption B: ...\textbackslash nOption C: ...\textbackslash nOption D: ...Option E: ...\textbackslash nOption F: Can produce simple, mainly isolated phrases about people and places. Please select the letter corresponding to the most appropriate option, and only output that letter without any additional comments or text.\textbackslash n\textbackslash nResult is Option:''} \footnote{To save space, we have omitted some text in this section, please refer to page 58 of \cite{council2001common} for the descriptors of all levels.}



\section{Experiments}

\subsection{Data Setup}

\begin{table}[!htbp]
    \centering
    \footnotesize
    \caption{The number of test submissions and hours of speech data for each corpus.}
    \begin{tabular}{c|cccc|cc}
    \toprule
    Corpus & \multicolumn{4}{c|}{Linguaskill} &  \multicolumn{2}{c}{S\&I} \\
    Split & Train & Dev & LinGen & LinBus & Train & Dev \\
    \midrule
    \#Sub. & 31,476 & 1,033 & 1,049 & 712 & 6,640 & 438 \\
    Hours & 3,465 & 114 & 115 & 78 & 244 & 35 \\
    \bottomrule
    \end{tabular}
    \label{tab:datasets}
\end{table}

This paper uses two datasets for building and evaluating spoken language assessment systems: a private dataset, Lingualskill \cite{ludlow2020official}; and a public corpus, Speak \& Improve Corpus 2025 (S\&I) \cite{knill2024speak}. Each test from Linguaskill and S\&I contains five parts focusing on different aspects of the candidate's language proficiency: For Part 1, the candidate is prompted to answer eight questions, of which the first two responses are not marked. The first four answers last 10 seconds, and the last four last 20 seconds. The second part is a read-aloud test where the candidate is instructed to read 8 sentences, each sentence is about 10 seconds long.
In parts 3 and 4, the candidate is instructed to talk for 1 minute to give their opinion on a given topic or describe the process depicted in the given diagram. Part 5 prompts the learner to answer 5 questions related to a topic, each question is around 20 seconds long.
Each part is scored from 1.0 to 6.0, roughly corresponding to the CEFR levels A1 to C2, and the average score of 5 parts is used as the candidate's overall score. Linguaskill provides two test sets LinGen and LinBus, which focus on General English and Business English scenarios, respectively. For S\&I, a development set has been pre-released as part of the S\&I Challenge 2025 \cite{qian2024speak} (parts 1,3-5 only). In the experiments, we report calibrated SLA results, where the development set is used to extract linear calibration factors.

\subsection{Model Setup}
In this paper, we conduct experiments using Qwen2-Audio-7B-Instruct, a model that has demonstrated outstanding performance across a range of audio understanding tasks \cite{chu2024qwen2}. In the following, we refer to this model as \textbf{Qwen2Audio}. In the experiments, we evaluate the zero-shot classification performance of Qwen2Audio on L2 oral assessment. When adapted, LoRA adaptors with a rank of 16 are inserted into both encoder and decoder layers, adding 10M parameters to the pre-trained model. 
In all settings, the speech LLM is trained for 2 epochs on the provided training set. For models trained on Linguaskill, a training batch size of 64 is used, while for S\&I, a batch size of 8 is employed. In the training, the learning rate is set to 1e-4 and a cosine learning rate scheduling strategy with weight decay is applied. Since Qwen2Audio only accepts audio inputs of less than 30 seconds, for parts 3 and 4, the original audio is split into two 30-second chunks in the experiments. We train the model to predict the reference score given each chunk, and during evaluation, we calculate the average score across both segments.

To evaluate the effectiveness of the grading systems, we report the model performance of several metrics on the test sets: root mean squared error (RMSE), Pearson's correlation coefficient (PCC), and Spearman's rank correlation coefficient (SRC).
For RMSE, a lower value indicates better system performance, whereas for other metrics, higher values are desirable. 
Several previous state-of-the-art grading systems are compared with our proposed approach. 
\cite{bannoassessment} builds a novel assessment system from the wav2vec 2.0 system and shows promising performance. 
The system uses a mean pooling approach to aggregate the input sequence. By replacing the mean pooling with an attention mechanism, similar to \cite{mcknight23_slate}, the grader achieves enhanced performance. 
This baseline model is referred to as the \textbf{wav2vec2} grader in the experiments. 
A cascaded \textbf{BERT} grader is also compared, where the ASR hypotheses of the response are used as input to train the BERT model \cite{raina20_interspeech, bannoassessment}. Here, the underlying ASR system is the Whisper small.en model \cite{radford2023robust}.

\begin{table}[!htbp]
    \centering
    \footnotesize
    \renewcommand\tabcolsep{3.9pt}
    \caption{L2 grading performance on part 5 of LinGen and LinBus using AudioQwen2. Models are trained on part 5 training data from Linguaskill and calibrated on the dev set.}
    \begin{tabular}{ll|ccc|ccc}
    \toprule
        \multirow{2}*{Train} & \multirow{2}*{Decode} & \multicolumn{3}{c|}{LinGen} & \multicolumn{3}{c}{LinBus} \\
        && RMSE & PCC & SRC & RMSE & PCC & SRC \\
        \midrule
        \multirow{2}*{${\tt zs}$} & hard & 1.227 & 0.178 & 0.167 & 1.246 & 0.394 & 0.397 \\
        & soft & 1.147 & 0.371 & 0.371 & 1.167 & 0.507 & 0.511 \\
        \midrule
        \multirow{2}*{${\tt ce}$} & hard & 0.615 & 0.879 & 0.882 & 0.663 & 0.865 & 0.869 \\
        & soft & 0.563 & 0.890 & 0.890 & 0.594 & 0.882 & 0.884 \\
        \midrule
        \multirow{2}*{${\tt fa}$} & hard & 0.592 & 0.877 & 0.881 & 0.604 & 0.876 & 0.877 \\
        & soft & \textbf{0.559} & \textbf{0.892} & \textbf{0.892} & \textbf{0.589} & \textbf{0.885} & \textbf{0.886} \\
        \midrule
       \multirow{1}*{${\tt reg}$} & - & 0.565 & 0.890 & 0.890 & 0.591 & 0.884 & 0.885 \\
        \bottomrule
    \end{tabular}
    \label{tab:compare}
\end{table}

\subsection{Experiments on L2 Grading}

\noindent
In Table \ref{tab:compare}, we compare the proposed training and evaluation strategies on the part 5 test of Linguaskill. For experiments denoted with ${\tt zs}$, we examine the capability of Qwen2Audio for assessing L2 speaking proficiency without additional training.
Influenced by the implicit positional bias of LLM \cite{liusie2023mitigating}, the majority of Qwen2Audio's outputs are \textit{``C''} and \textit{``D''}, leading to poor performance on the test sets. 
With the \textit{soft} decoding mode, we compute a fair average of the predicted score rather than only considering the highest model prediction, resulting in improved performance. 
We also list the performance of models adapted with LoRA tuning.
By only introducing 10M new parameters, the model performance on all metrics largely improve. After the training, Qwen2Audio learns to more faithfully grade the learner's performance, with the PCC value increasing from 0.371 to 0.892 on LinGen. 
For models trained with different losses, the proposed ${\tt fa}$ approach shows the best performance. Compared to training with the ${\tt ce}$ target, this practice represents the scores more accurately, especially for the .5 annotations. 
Additionally, when compared to the regression model, it is more aligned with the pre-training task of LLMs to generate text tokens, leading to superior performance. Results on the dev set show that the models trained with ${\tt ce}$ or ${\tt fa}$ losses converge faster than the regression model, indicating a smoother training process. Since the ${\tt fa}$ model with \textit{soft} decoding setup shows the best performance, we present the Qwen2Audio results using this setup in the following experiments, unless stated otherwise.


\begin{table}[!htbp]
    \footnotesize
    \renewcommand\tabcolsep{4pt}
    \centering
    \caption{Submission-level evaluation for graders trained on five parts of Linguaskill. Results are calibrated on the dev set.}
    \begin{tabular}{l|ccc|ccc}
    \toprule
    \multirow{2}*{Model} & \multicolumn{3}{c}{LinGen} & \multicolumn{3}{c}{LinBus} \\
    & RMSE & PCC & SRC & RMSE & PCC & SRC \\
    \midrule
    wav2vec2 & 0.383 & 0.934 & 0.938 & 0.380 & 0.922 & 0.923 \\
    BERT & 0.356 & 0.942 & 0.948 & 0.381 & 0.928 & 0.934 \\
    Qwen2Audio-${\tt zs}$ & 0.807 & 0.640 & 0.662 & 0.894 & 0.639 & 0.661 \\
    Qwen2Audio-${\tt fa}$ & \textbf{0.323} & \textbf{0.954} & \textbf{0.958} & \textbf{0.356} & \textbf{0.938} & \textbf{0.940} \\
    \bottomrule
    \end{tabular}
    \label{tab:overall}
\end{table}

\begin{figure}[H]
    \centering
    \includegraphics[width=0.495\linewidth]{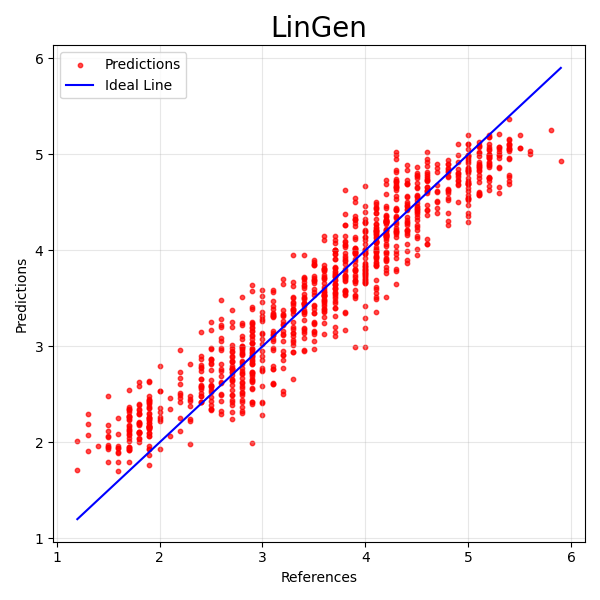}
    \includegraphics[width=0.495\linewidth]{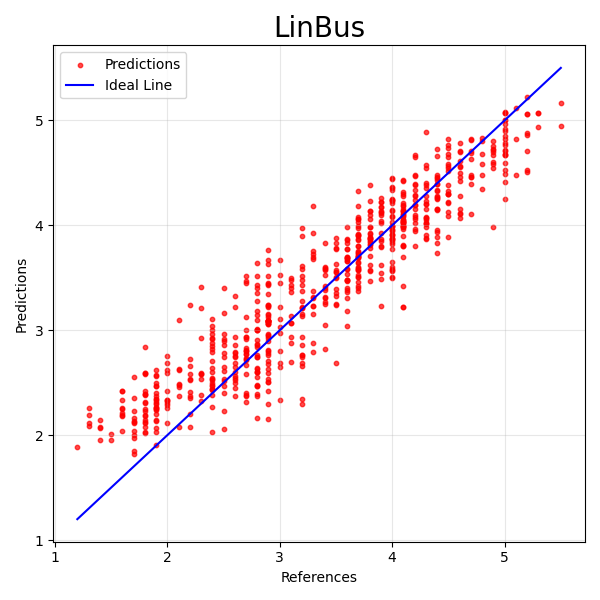}
    \caption{References vs predictions using Audio2Qwen graders.}
    \label{fig:dot}
\end{figure}

In Table \ref{tab:overall}, we show the overall assessment performance across all parts on Linguaskill. Here, one grader is trained for each part of the test submission, with the final score calculated as the average scores from all 5 parts. Given the strong performance of the baseline graders on this task, achieving further improvements is inherently challenging.
Despite this, our Qwen2Audio-${\tt fa}$ grader is able to achieve considerate advancements, facilitated by the audio understanding knowledge gained during pre-training.
Even the graders built in the zero-shot setting achieve overall PCCs of 0.640 and 0.639, indicating the potential emergent ability of speech LLMs in L2 spoken language assessment.
Moreover, the proposed approach directly predicts scores based on the input utterance, eliminating the need for ASR decoding in the BERT grader. This prevents information loss that occurs in a cascaded system and reduces the computational cost associated with ASR decoding. 
The predictions against reference scores are depicted in Figure \ref{fig:dot}, showcasing a high correlation between the automatic grader's predictions and the human annotations, which highlights the great potential of building auto graders using speech LLMs.

%


\subsection{Analysis on Generalisation Ability}
\begin{table}[!htbp]
    \centering
    \footnotesize
    \caption{Effect on PCC of matched and cross-part grader models for LinGen.}
    \renewcommand\tabcolsep{3pt}
    \begin{tabular}{@{}c|l|ccccc|c@{}}
    \toprule
    Setup & Model & P1 & P2 & P3 & P4 & P5 & Overall \\
    \midrule
    \multirow{3}*{matched}
    & wav2vec2 & 0.844 & 0.834 & 0.841 & 0.818 & 0.874 & 0.934 \\
    & BERT & 0.854 & 0.771 & 0.861 & 0.838 & 0.884 & 0.942 \\ 
    & Qwen2Audio & \textbf{0.870} & \textbf{0.845} & \textbf{0.874} & \textbf{0.861} & \textbf{0.892} & \textbf{0.954} \\
    \midrule
    read
    & wav2vec2 & 0.672 & 0.834 & 0.631 & 0.680 & 0.748 & 0.847   \\
    aloud & BERT & 0.763 & 0.771 & 0.689 & 0.701 & 0.767 & 0.888 \\ 
    (P2) & Qwen2Audio & \textbf{0.772} & \textbf{0.845} & \textbf{0.767} & \textbf{0.807} & \textbf{0.848} & \textbf{0.911} \\
    \midrule
 short
    & wav2vec2 & 0.824 & 0.676 & 0.835 & 0.815 & 0.874 & 0.921 \\
    open & BERT & 0.843 & 0.644 & 0.830 & 0.821 & 0.884 & 0.936\\
    (P5) & Qwen2Audio & \textbf{0.855} & \textbf{0.816} & \textbf{0.864} & \textbf{0.855} & \textbf{0.892} & \textbf{0.950} \\
    \bottomrule
    \end{tabular}
    \label{tab:transfer}
\end{table}

\noindent
One advantage of LLMs is their strong generalisation capabilities, enabling them to perform well on tasks that they are not specifically trained on. In this section, we aim to assess the generalisation capability of the grader on out-of-domain test sets. For Linguaskill, although all five parts of the test aim to measure the candidate's spoken language proficiency, they vary in format and focus. For instance, part 2 requires the student to read aloud the given sentences while other parts feature spontaneous answers. Moreover, parts 1 and 5 consist of short responses (10 to 20 seconds), while parts 3 and 4 contain longer utterances. Table 6 evaluates the cross-part performance of the trained graders. In the \textit{matched} setup, graders are trained and evaluated separately on the data from each part of Linguaskill tests, consistent with the settings in Table \ref{tab:overall}.
Furthermore, we evaluate the cross-part transferability of the trained graders by applying the model trained solely on part 2 or part 5 to the test data from all parts. In this setting, we aim to assess whether models trained on read-aloud tests can effectively generalise to free-speaking tests, and vice versa.

Surprisingly, only using the read aloud data from part 2, we are able to build an effective grader that has an overall PCC of 0.911. The results suggest that the trained Qwen2Audio is able to assess the candidate's performance in terms of high-level features such as grammar and word usage. In contrast, wav2vec2 shows degraded performance in this scenario. For the BERT grader, it even fails on the part 2 test as information of the audio is lost in the cascaded pipeline by only passing the ASR outputs. The result is promising and indicates a broader application of the proposed method, as the read-aloud corpora are easier to collect and have greater availability compared to spontaneous speech datasets. 
Furthermore, the Qwen2Audio model trained solely on part 5 demonstrates excellent performance in the transfer setup, even on the part 2 test, and outperforms the wav2vec2 and BERT baselines from the matched case.

Another form of generalisation is the ability to apply models to a different task without the need of retraining. In Table \ref{tab:crosstask}, we compare the system performance of 
graders trained on Linguaskill and on Speak and Improve (S\&I) on the Linguaskill test sets, LinGen and LinBus, and the S\&I dev set. As the part 2 data is not released in the S\&I corpus, we evaluate the overall system performance on the other 4 parts for all test sets.
Graders trained on S\&I data illustrate the model’s performance when training data is limited.
In comparison with the baseline BERT grader, Qwen2Audio systems achieve a substantial performance improvement on the PCC metric (0.833 over 0.753), demonstrating its effectiveness in the low-resource setting. Additionally, we list the results of the cross-task evaluation. Compared to the BERT grader, Qwen2Audio showcases remarkable generalisation ability in this scenario where models trained on Linguaskill are evaluated on the S\&I dev set and models trained on S\&I are tested on Linguaskill test sets. These experiments highlight the robustness of the proposed L2 graders.

\begin{table}[!htbp]
    \centering
    \caption{Within and cross task evaluation on test parts 1,3,4,5 using models trained on Linguaskill and S\&I.}
    \footnotesize
    \begin{tabular}{c|c|c|c|c}
    \toprule
        \multirow{2}*{Train} & \multirow{2}*{Model} & \multicolumn{3}{c}{Test (PCC)} \\
         & & LinGen & LinBus & S\&I \\
         \midrule
         \multirow{2}*{LNG} & BERT & 0.941 & 0.931 & 0.796 \\ 
         & Qwen2Audio & \textbf{0.951} & \textbf{0.938} & \textbf{0.824} \\
        \midrule
        \multirow{2}*{S\&I} & BERT & 0.901 & 0.901 & 0.753  \\
         & Qwen2Audio & \textbf{0.929} & \textbf{0.914} & \textbf{0.833} \\
        \bottomrule
    \end{tabular}
    \label{tab:crosstask}
\end{table}


\section{Conclusions}
In this paper, we examine the effectiveness of building spoken language assessment systems using speech LLMs. Various training and decoding schemes are compared on Linguaskill and S\&I datasets. The zero-shot experiments show the potential of leveraging speech LLMs for L2 oral scoring. With further adaptation, the Qwen2Audio grader trained with fair average loss achieves the best overall performance on both general and business test sets, reaching PCC of 0.954 and 0.938, respectively. The proposed approach overperforms strong baselines of the BERT-based cascaded system and an E2E system built from wav2vec 2.0.  Experiments on S\&I further validate the effectiveness of our proposed methods when limited training data is available.
Additionally, speech LLM graders show great generalisation capability when applied to a different test part or a different dataset.
In our future work, we aim to further explore the model's emergent capabilities for L2 scoring, such as providing feedback on the reasoning behind its assigned scores.



\clearpage
\bibliographystyle{IEEEtran}
\bibliography{mybib}

\begin{thebibliography}{10}
\providecommand{\url}[1]{#1}
\csname url@samestyle\endcsname
\providecommand{\newblock}{\relax}
\providecommand{\bibinfo}[2]{#2}
\providecommand{\BIBentrySTDinterwordspacing}{\spaceskip=0pt\relax}
\providecommand{\BIBentryALTinterwordstretchfactor}{4}
\providecommand{\BIBentryALTinterwordspacing}{\spaceskip=\fontdimen2\font plus
\BIBentryALTinterwordstretchfactor\fontdimen3\font minus \fontdimen4\font\relax}
\providecommand{\BIBforeignlanguage}[2]{{%
\expandafter\ifx\csname l@#1\endcsname\relax
\typeout{** WARNING: IEEEtran.bst: No hyphenation pattern has been}%
\typeout{** loaded for the language `#1'. Using the pattern for}%
\typeout{** the default language instead.}%
\else
\language=\csname l@#1\endcsname
\fi
#2}}
\providecommand{\BIBdecl}{\relax}
\BIBdecl

\bibitem{bernstein1990automatic}
J.~Bernstein, M.~Cohen, H.~Murveit, D.~Rtischev, and M.~Weintraub, ``Automatic evaluation and training in {English} pronunciation,'' in \emph{{First International Conference on Spoken Language Processing}}, 1990.

\bibitem{cucchiarini1997automatic}
C.~Cucchiarini, H.~Strik, and L.~Boves, ``Automatic evaluation of {Dutch} pronunciation by using speech recognition technology,'' in \emph{{1997 IEEE Workshop on Automatic Speech Recognition and Understanding Proceedings}}, 1997, pp. 622--629.

\bibitem{franco2000sri}
H.~Franco, V.~Abrash, K.~Precoda, H.~Bratt, R.~Rao, J.~Butzberger, R.~Rossier, and F.~Cesari, ``The {SRI EduSpeak\texttrademark} system: Recognition and pronunciation scoring for language learning,'' in \emph{Proceedings of {InSTILL}}, 2000, pp. 123--128.

\bibitem{zechner2007speechrater}
K.~Zechner, D.~Higgins, and X.~Xi, ``Speechrater: A construct-driven approach to scoring spontaneous non-native speech,'' in \emph{Proc. SLaTE}, 2007, pp. 128--131.

\bibitem{qian2012use}
X.~Qian, H.~Meng, and F.~K. Soong, ``The use of {DBN-HMMs} for mispronunciation detection and diagnosis in {L2 English} to support computer-aided pronunciation training,'' in \emph{Proc. Interspeech}, 2012.

\bibitem{evanini2018improvements}
K.~Evanini, M.~Mulholland, R.~Ubale, Y.~Qian, R.~A. Pugh, V.~Ramanarayanan, and A.~Cahill, ``Improvements to an automated content scoring system for {Spoken CALL} responses: the {ETS} submission to the {Second Spoken CALL Shared Task}.'' in \emph{{Proc. Interspeech}}, 2018, pp. 2379--2383.

\bibitem{qian2019slate}
M.~Qian, P.~Jančovič, and M.~Russell, ``{The University of Birmingham 2019 Spoken CALL Shared Task Systems: Exploring the importance of word order in text processing},'' in \emph{8th ISCA Workshop on Speech and Language Technology in Education (SLaTE 2019)}, 2019, pp. 11--15.

\bibitem{devlin2019}
J.~Devlin, M.-W. Chang, K.~Lee, and K.~Toutanova, ``{BERT}: Pre-training of deep bidirectional transformers for language understanding,'' in \emph{{Proceedings of the 2019 Conference of the North American Chapter of the Association for Computational Linguistics: Human Language Technologies, Volume 1 (Long and Short Papers)}}, 2019, pp. 4171--4186.

\bibitem{raina2020universal}
V.~Raina, M.~J.~F. Gales, and K.~M. Knill, ``Universal adversarial attacks on spoken language assessment systems,'' in \emph{{Proc. Interspeech}}, 2020, pp. 3855--3859.

\bibitem{wang2021}
X.~Wang, K.~Evanini, Y.~Qian, and M.~Mulholland, ``Automated scoring of spontaneous speech from young learners of {English} using transformers,'' in \emph{{2021 IEEE Spoken Language Technology Workshop (SLT)}}, 2021, pp. 705--712.

\bibitem{baevski}
A.~Baevski, H.~Zhou, A.~Mohamed, and M.~Auli, ``wav2vec 2.0: A framework for self-supervised learning of speech representations,'' in \emph{{Proceedings of the 34th Conference on Neural Information Processing Systems (NeurIPS 2020)}}, 2020, pp. 1--12.

\bibitem{Hsu2021HuBERTSS}
W.-N. Hsu, B.~Bolte, Y.-H.~H. Tsai, K.~Lakhotia, R.~Salakhutdinov, and A.~Mohamed, ``{HuBERT}: Self-supervised speech representation learning by masked prediction of hidden units,'' \emph{IEEE/ACM Transactions on Audio, Speech, and Language Processing}, vol.~29, pp. 3451--3460, 2021.

\bibitem{peng21}
L.~Peng, K.~Fu, B.~Lin, D.~Ke, and J.~Zhan, ``{A Study on Fine-Tuning wav2vec2.0 Model for the Task of Mispronunciation Detection and Diagnosis},'' in \emph{Proc. Interspeech}, 2021, pp. 4448--4452.

\bibitem{wu21}
M.~Wu, K.~Li, {W.-K. Leung}, and H.~Meng, ``Transformer based end-to-end mispronunciation detection and diagnosis,'' in \emph{{Proc. Interspeech}}, 2021, pp. 3954--3958.

\bibitem{xu21}
X.~Xu, Y.~Kang, S.~Cao, B.~Lin, and L.~Ma, ``{Explore wav2vec 2.0 for Mispronunciation Detection},'' in \emph{{Proc. Interspeech}}, 2021, pp. 4428--4432.

\bibitem{eesung}
E.~{Kim}, J.-J. {Jeon}, H.~{Seo}, and H.~{Kim}, ``{Automatic Pronunciation Assessment using Self-Supervised Speech Representation Learning},'' in \emph{Proc. Interspeech}, 2022, pp. 1411--1415.

\bibitem{banno2023proficiency}
S.~Bann{\`o} and M.~Matassoni, ``{Proficiency assessment of L2 spoken English using wav2vec 2.0},'' in \emph{2022 IEEE Spoken Language Technology Workshop (SLT)}.\hskip 1em plus 0.5em minus 0.4em\relax IEEE, 2023, pp. 1088--1095.

\bibitem{bannoassessment}
S.~Bann{\`o}, K.~M. Knill, M.~Matassoni, V.~Raina, and M.~Gales, ``{Assessment of L2 Oral Proficiency Using Self-Supervised Speech Representation Learning},'' in \emph{Proc. 9th Workshop on Speech and Language Technology in Education (SLaTE)}, 2023, pp. 126--130.

\bibitem{mcknight23_slate}
S.~W. McKnight, A.~Civelekoglu, M.~Gales, S.~Bannò, A.~Liusie, and K.~M. Knill, ``Automatic assessment of conversational speaking tests,'' in \emph{9th Workshop on Speech and Language Technology in Education (SLaTE)}, 2023, pp. 99--103.

\bibitem{tangsalmonn}
C.~Tang, W.~Yu, G.~Sun, X.~Chen, T.~Tan, W.~Li, L.~Lu, M.~Zejun, and C.~Zhang, ``{SALMONN: Towards Generic Hearing Abilities for Large Language Models},'' in \emph{The Twelfth International Conference on Learning Representations}.

\bibitem{chu2023qwen}
Y.~Chu, J.~Xu, X.~Zhou, Q.~Yang, S.~Zhang, Z.~Yan, C.~Zhou, and J.~Zhou, ``{Qwen-audio: Advancing universal audio understanding via unified large-scale audio-language models},'' \emph{arXiv preprint arXiv:2311.07919}, 2023.

\bibitem{chu2024qwen2}
Y.~Chu, J.~Xu, Q.~Yang, H.~Wei, X.~Wei, Z.~Guo, Y.~Leng, Y.~Lv, J.~He, J.~Lin \emph{et~al.}, ``{Qwen2-audio technical report},'' \emph{arXiv preprint arXiv:2407.10759}, 2024.

\bibitem{mizumoto2023gpt}
A.~Mizumoto and M.~Eguchi, ``Exploring the potential of using an ai language model for automated essay scoring,'' \emph{Research Methods in Applied Linguistics}, vol.~2, no.~2, p. 100050, 2023.

\bibitem{yancey-etal-2023-rating}
K.~P. Yancey, G.~Laflair, A.~Verardi, and J.~Burstein, ``Rating short {L}2 essays on the {CEFR} scale with {GPT}-4,'' in \emph{Proceedings of the 18th Workshop on Innovative Use of NLP for Building Educational Applications (BEA 2023)}.\hskip 1em plus 0.5em minus 0.4em\relax Toronto, Canada: Association for Computational Linguistics, Jul. 2023, pp. 576--584.

\bibitem{banno-etal-2024-gpt}
S.~Bannò, H.~K. Vydana, K.~Knill, and M.~Gales, ``Can {GPT}-4 do {L}2 analytic assessment?'' in \emph{Proceedings of the 19th Workshop on Innovative Use of NLP for Building Educational Applications (BEA 2024)}.\hskip 1em plus 0.5em minus 0.4em\relax Mexico City, Mexico: Association for Computational Linguistics, Jun. 2024, pp. 149--164.

\bibitem{fu2024pronunciation}
K.~Fu, L.~Peng, N.~Yang, and S.~Zhou, ``{Pronunciation Assessment with Multi-modal Large Language Models},'' \emph{arXiv preprint arXiv:2407.09209}, 2024.

\bibitem{wei2022emergent}
J.~Wei, Y.~Tay, R.~Bommasani, C.~Raffel, B.~Zoph, S.~Borgeaud, D.~Yogatama, M.~Bosma, D.~Zhou, D.~Metzler \emph{et~al.}, ``{Emergent Abilities of Large Language Models},'' \emph{Transactions on Machine Learning Research}, 2022.

\bibitem{council2001common}
C.~of~Europe. Council for Cultural Co-operation. Education Committee. Modern Languages~Division, \emph{{Common European framework of reference for languages: Learning, teaching, assessment}}.\hskip 1em plus 0.5em minus 0.4em\relax Cambridge University Press, 2001.

\bibitem{ludlow2020official}
K.~Ludlow, \emph{{Official Quick Guide to Linguaskill}}.\hskip 1em plus 0.5em minus 0.4em\relax Cambridge University Press, 2020.

\bibitem{knill2024speak}
\BIBentryALTinterwordspacing
K.~Knill, D.~Nicholls, M.~J. Gales, M.~Qian, and P.~Stroinski, ``{The Speak \& Improve Corpus 2025: an L2 English Speech Corpus for Language Assessment and Feedback},'' 2025. [Online]. Available: \url{https://doi.org/10.17863/CAM.114333}
\BIBentrySTDinterwordspacing

\bibitem{qian2024speak}
M.~Qian, K.~Knill, S.~Banno, S.~Tang, P.~Karanasou, M.~J. Gales, and D.~Nicholls, ``{Speak \& Improve Challenge 2025: Tasks and Baseline Systems},'' \emph{arXiv preprint arXiv:2412.11985}, 2024.

\bibitem{raina20_interspeech}
V.~Raina, M.~J. Gales, and K.~M. Knill, ``Universal adversarial attacks on spoken language assessment systems,'' in \emph{Proc. Interspeech}, 2020, pp. 3855--3859.

\bibitem{radford2023robust}
A.~Radford, J.~W. Kim, T.~Xu, G.~Brockman, C.~McLeavey, and I.~Sutskever, ``Robust speech recognition via large-scale weak supervision,'' in \emph{International conference on machine learning}.\hskip 1em plus 0.5em minus 0.4em\relax PMLR, 2023, pp. 28\,492--28\,518.

\bibitem{liusie2023mitigating}
A.~Liusie, P.~Manakul, and M.~Gales, ``{Mitigating Word Bias in Zero-shot Prompt-based Classifiers},'' in \emph{Findings of the Association for Computational Linguistics: IJCNLP-AACL 2023 (Findings)}, 2023, pp. 327--335.

\end{thebibliography}

\end{document}